\documentclass[
]{ceurart}

\usepackage{listings}
\usepackage{xcolor, soul}
\sethlcolor{yellow}

\begin{document}

\copyrightyear{2023}
\copyrightclause{Copyright for this paper by its authors.
  Use permitted under Creative Commons License Attribution 4.0
  International (CC BY 4.0).}

\conference{Empowering Education with LLMs -- the Next-Gen Interface and Content Generation}

\title{On the application of Large Language Models for language teaching and assessment technology}

%\tnotetext[1]{You can use this document as the template for preparing your publication. We recommend using the latest version of the ceurart style.}


\author[1]{Andrew Caines}[%
email=andrew.caines@cl.cam.ac.uk
]
%\cormark[1]
\address[1]{ALTA Institute \& Computer Laboratory, University of Cambridge}

\author[1]{Luca Benedetto}[%
email=luca.benedetto@cl.cam.ac.uk
]
\author[1]{Shiva Taslimipoor}
\author[1]{Christopher Davis}
\author[1]{Yuan Gao}
\author[1]{{\O}istein Andersen}

\author[2,1]{Zheng Yuan}
\address[2]{King's College London}

\author[3,1]{Mark Elliott}
\address[3]{Cambridge University Press \& Assessment}

\author[1]{Russell Moore}

\author[4,1]{Christopher Bryant}
\address[4]{Writer, Inc.}

\author[5,1]{Marek Rei}
\address[5]{Imperial College London}

\author[2,1]{Helen Yannakoudakis}

\author[3]{Andrew Mullooly}

\author[6]{Diane Nicholls}
\address[6]{English Language iTutoring (ELiT)}

\author[1]{Paula Buttery}[%
email=paula.buttery@cl.cam.ac.uk
]

%\cortext[1]{Corresponding author.}
%\fntext[1]{These authors contributed equally.}

\begin{abstract}
The recent release of very large language models such as PaLM and GPT-4 has made an unprecedented impact in the popular media and public consciousness, giving rise to a mixture of excitement and fear as to their capabilities and potential uses, and shining a light on natural language processing research which had not previously received so much attention. The developments offer great promise for education technology, and in this paper we look specifically at the potential for incorporating large language models in AI-driven language teaching and assessment systems. We consider several research areas -- content creation and calibration, assessment and feedback -- and also discuss the risks and ethical considerations surrounding generative AI in education technology for language learners. Overall we find that larger language models offer improvements over previous models in text generation, opening up routes toward content generation which had not previously been plausible. For text generation they must be prompted carefully and their outputs may need to be reshaped before they are ready for use.
%-- therefore human experts must still be in the loop. 
For automated grading and grammatical error correction, tasks whose progress is checked on well-known benchmarks, early investigations indicate that large language models on their own do not improve on state-of-the-art results according to standard evaluation metrics. For grading it appears that linguistic features established in the literature should still be used for best performance, and for error correction it may be that the models can offer alternative feedback styles which are not measured sensitively with existing methods. In all cases, there is work to be done to experiment with the inclusion of large language models in education technology for language learners, in order to properly understand and report on their capacities and limitations, and to ensure that foreseeable risks such as misinformation and harmful bias are mitigated.
\end{abstract}


\maketitle

\section{Introduction}

The training of \emph{large language models} (LLMs) -- also known as \emph{pre-trained language models} or \emph{foundation models} -- has had a transformative effect on the fields of natural language processing (NLP) and artificial intelligence (AI) more broadly. LLMs are `large' because they are neural networks made up of billions or trillions of parameters. The networks are Transformers \cite{vaswani} trained on huge swathes of text from the World Wide Web, using language modelling objectives such as predicting omitted (or, `masked') words and sentence pairs \cite{bert}, or predicting the next token in a sequence \cite{radford-gpt1}. Furthermore, in the few-shot learning paradigm, LLMs can be directed towards new tasks without large quantities of task-specific data \cite{radford2019language,brown-gpt3}, the collection of which tends to be time-consuming and costly. Overall, LLMs also offer great potential for educational applications. One previous paper has already provided an overview of some of the possible applications of LLMs to educational technology as a whole \cite{kasneci2023chatgpt}, across subjects. Our distinct contribution is to focus on the language learning and assessment domain specifically. In this paper, we describe some of the uses for LLMs in the context of language learning, discuss the state of the art or work in progress, and consider practical, societal and ethical implications.

We set out a number of uses for LLMs in the language learning domain, relating to content creation and calibration, automated assessment of written texts, and personalised feedback. In each case the general principles of the approach are well established thanks to previous work with pre-existing LLMs such as BERT \cite{bert} -- language models with millions of parameters have existed for several years already. We look at the opportunities presented by the recent and rapid steps taken by OpenAI in releasing new variants from the `generative pre-training' (GPT) model series, along with some newly published pre-prints relating to LLMs and the language learning research field. We refer to some LLM-driven language learning applications already in use, and outline the variety of LLMs available besides GPT. It is a fast evolving research field, one being driven by industry developments. We perceive some possible risks in this research trajectory, which include but are not limited to the absence of proper safeguards on education technology, the lack of public understanding as to how LLMs are trained and how they can confidently assert incorrect information, and the harm to the advancement of education technology as a whole if it is considered `solved' by investors and research councils -- not to mention the ethical issues that are already well known, such as data protection \cite{carlini2021extracting}, examination malpractice \cite{fyfe2022cheat, ventayen2023openai}\footnote{Note that the text-matching tool \texttt{Turnitin} \cite{Mphahlele2019}, which is commonly used to detect plagiarism, has developed a module to detect the use of AI in essays: \url{https://www.turnitin.com/blog/the-launch-of-turnitins-ai-writing-detector-and-the-road-ahead}}, environmental impact \cite{strubell-etal-2019-energy,dodge2022}, and internet addiction \cite{chou2005review,bhargava2021ethics,gioia2021problematic}, among others.

\section{Large Language Models \& Language Learning EdTech}

At the time of writing, one of the most prominent LLMs is OpenAI's GPT-4 \cite{openai2023gpt4} released in March 2023 after six months of pre-launch work improving safety and fact-checking, also for product development with selected partners including the education technology (EdTech) firms Duolingo\footnote{\url{https://blog.duolingo.com/duolingo-max/}} and Khan Academy\footnote{\url{https://blog.khanacademy.org/harnessing-ai-so-that-all-students-benefit-a-nonprofit-approach-for-equal-access/}}. This built on the prior success and notoriety of GPT-3, released in June 2020, along with its related chatbot application, \texttt{ChatGPT}\footnote{\url{https://openai.com/blog/chatgpt}}. 
Recently, we have seen more focus on the effects and implications of using chatbots for creating interactions with language learners \cite{huang2022,tyen-etal-2022-towards}.

Of most relevance here is the partnership between OpenAI and Duolingo, the language learning application developer, which resulted in the subscription service \texttt{Duolingo Max} \cite{duolingo-max}. Duolingo Max presents two new features: `Role Play' and `Explain My Answer'. The former involves some limited conversation towards a goal such as ordering food, which starts with a pre-scripted prompt but then proceeds over several open-ended chat turns between user and chatbot. The latter is an option for additional feedback on grammatical points, involving a limited dialogue with pre-specified responses for the user to guide the conversation (e.g.\ ``Yes, I'm all set'', ``Can I see an example?'', ``No, please elaborate''). Another limitation is that the service is only currently available in selected countries and for a few languages.

Nevertheless, this development points towards further opportunities in AI-driven education technology for language learning, as discussed below. It should be noted that there are many alternatives to the GPT models, including the `text-to-text Transformer' (T5) \cite{t5}, PaLM (Parallel Language Model) \cite{palm2} and LaMDA (Language Model for Dialogue Applications) \cite{lamda} by Google; LLaMA \cite{llama} and Open Pre-trained Transformers (OPT) \cite{opt} by Meta AI; and DeepMind's Gopher \cite{gopher}.
At least a few of these models are `multilingual', having been trained on corpora from multiple languages, albeit with a strong bias towards English\footnote{e.g.\ See the distribution of languages in the training data for GPT-3: \url{https://github.com/openai/gpt-3/blob/master/dataset_statistics/languages_by_document_count.csv}}. In addition there are models which have been trained on bilingual data, notably Chinese--English \cite{zeng2022glm130b} and Russian--English\footnote{\url{https://github.com/yandex/YaLM-100B}}.
Alongside LLM developments by large technology companies, we also note the various open-source efforts to train on known datasets (e.g.\ EleutherAI's GPT-X \cite{black-etal-2022-gpt} and \emph{The Pile} \cite{gao2020pile}), or as massive research collaborations (e.g.\ BLOOM: the BigScience Large Open-science Open-access Multilingual Language Model \cite{workshop2023bloom}), or to democratise LLMs for wider use in web applications involving natural language interfaces (e.g.\ \texttt{langchain}\footnote{\url{https://python.langchain.com/en/latest/index.html}}, Cohere AI\footnote{\url{https://cohere.com/}}, and Transformers \cite{wolf-etal-2020-transformers}). There are also open-source alternatives to ChatGPT, such as Open Assistant\footnote{\url{https://open-assistant.io/}} and StableVicuna\footnote{\url{https://stability.ai/blog/stablevicuna-open-source-rlhf-chatbot}}.
Finally we highlight efforts to transparently evaluate LLMs in comprehensive and varied ways, for instance in the HELM project (Holistic Evaluation of Language Models) \cite{liang2022holistic}.

\section{Content Creation: Creating Assessment Items and Teaching Materials}

%\paragraph{Models}
Pre-trained transformer models are being explored to generate exam texts and items for educational purposes like in the Duolingo English Test \cite{ParkEtAl2022}, or in Google's quantitative reasoning application, Minerva \cite{lewkowycz2022solving}.
%, \textcolor{blue}{ or for text adaptation as in the TSAR-2022 Shared Task on Multilingual Lexical Simplification \cite{saggion-etal-2022-findings,aumiller-gertz-2022-unihd}}.
%While v
Variations of GPT models are best known to the wider public as being able to create common forms of language assessment tests. However, there are other successful pre-trained LLMs such as variations of BERT~\cite{bert}, BART~\cite{bart}, or T5~\cite{t5} which are popular among NLP scientists. Since these models are trained with different target task objectives, they should be prompted differently for various kinds of text generation. %However, 
None of them has been trained with a storytelling objective to generate fluent long texts as GPT* models have. Nevertheless, since the target tasks are better defined, evaluation of the performance of these models is more thorough and explainable. BERT-based models have shown impressive results in filling the gaps in sequences of text. BART achieves state-of-the-art results in generating parallel sentences as in machine translation or grammatical error correction~\cite{bart,katsumata-komachi-2020-stronger}. T5 modelled $24$ tasks as text-to-text generation and proved very successful in question answering and text summarisation~\cite{t5}.

These models are widely used for narrower tasks like question generation \cite{pan2019recent}, for reading comprehension exercises \cite{raina2022multiplechoice}, or prompt generation %\cite{AndrewCainesWorkOnSpeakingPrompts?} 
for writing and speaking. For example, Felice \emph{et al.}~\cite{felice-etal-2022-constructing} use BERT and ELECTRA \cite{electra} to predict the position of the gaps for designing high-quality cloze tests for language learners, while various pre-trained language models are used for generating distractors~\cite{chung-etal-2020-bert,chiang-etal-2022-cdgp,manakul2023mqag}.
%\textcolor{blue}{
In addition LLMs have been put to use for the purpose of text simplification, which is relevant for language learners in the context of reading comprehension exercises and adapting texts automatically to an appropriate level. Notably, a GPT-3 based solution by the UniHD team was the winning entry for the English track of the TSAR-2022 Shared Task on Multilingual Lexical Simplification \cite{saggion-etal-2022-findings,aumiller-gertz-2022-unihd}.
%}
%TODO

\paragraph{Datasets and Evaluation}
With the emergence of LLMs, large-scale datasets are required for evaluation.
Text generation methods are evaluated using text-similarity based metrics like BLEU \cite{bleu}, ROUGE \cite{rouge}, METEOR \cite{meteor}, and more recently BERTScore~\cite{zhang2020bertscore}, or learned evaluation metrics \cite{lowe2018automatic,shimanaka-etal-2018-ruse} which assess the correlation between generated texts (e.g.\ the question) and the ones originally written by human experts. Automatic evaluations require top-quality and expert-designed datasets. 
Available datasets for language learning exams for NLP research include RACE \cite{lai-etal-2017-race}, SCDE~\cite{kong2020scde} and CLOTH \cite{xie-etal-2018-large}. However, there are smaller-scale datasets such as CEPOC \cite{felice-etal-2022-cepoc} for Cloze test creation, and the Teacher-Student Chatroom Corpus \cite{tscc}, which can be used as test sets to evaluate zero-shot or few-shot learning models.  %AndrewC's teacher-student corpus?
Evaluation approaches for open-ended text generation are still far from being ideal.
Human-centric evaluations involve ranking the generated texts based on different factors, such as fluency and coherence of the generated texts, or its relevance to the context document and the answer (where available) \cite{hosking-riedel-2019}. In this new era of increasingly large language models, human evaluation is more difficult and time-consuming, leading researchers to design comparison datasets that contain human-labelled comparisons between outputs of different systems \cite{ouyang2022training}.

\paragraph{Human-in-the-loop content generation}
As an exploratory study, we have worked with publicly available GPT-3 models to generate open-ended texts and evaluate their suitability as a basis for low-stakes, self-study language learning exercises. Having a human-in-the-loop policy in mind, the prompts are engineered by a human expert, with post-generation text refinement and question authoring also carried out by experts. Such an approach can be seen as helping mitigate against various known risks associated with the output of LLMs (e.g., hallucinations, offensive content, stereotyping, etc).

All definable model parameters (Temperature, Frequency Penalty, Presence Penalty and Max Length) are kept to fixed levels throughout to limit the number of variables across the dataset. Input prompts containing target genre and key content points are designed by the human expert in order to provide a basis for possible testing foci at the target level. The key content points also help generate similar enough output texts from a single (or slightly modified) input prompt, to allow for collation of the best elements from multiple output versions.  

For this research, the generated texts are intended to support single B2 CEFR level\footnote{The Common European Framework of Reference for Languages (CEFR) organises language proficiency in six levels, A1 to C2, which represent Basic User, Independent User and Proficient User.} multiple choice reading comprehension questions with 3 answer options. The generated texts are reviewed by the human expert and given an `accept' or `reject' status based on their appropriateness for the target proficiency level and relevance to the content points. Accepted texts are added to a content pool, also containing fully human-authored texts. Another group of human experts (question writers) approach the accepted texts as they would any other content. For openness and transparency, question writers are informed in advance that the pool contains AI-generated content, but not which texts are AI-generated, and which have been written by human authors. Question writers select and edit the texts, writing one 3-option multiple choice question per text. The annotations of this dataset, including the accept/reject status and the measures of quality of the generated texts assessed by question writers (based on the amounts of edits made on the texts) can be used to train models which can automatically assess the generated texts in future. 
%When released, t
The dataset can also be used to train reward functions for further fine-tuning of the generative models.

ChatGPT has been trained using a combination of supervised fine-tuning and reinforcement learning from human feedback (RLHF)~\cite{ ziegler2020finetuning, stiennon2022learning}. It uses InstructGPT~\cite{ouyang2022training} and includes the steps: pre-training, fine-tuning, and reward learning. 
%Following the success of ChatGPT, which uses InstructGPT~\cite{ouyang2022training} and benefits from reinforcement learning from human feedback (RLHF)~\cite{ ziegler2020finetuning, stiennon2022learning}, we propose to use reinforcement learning to generate targeted output texts more successfully. In the development of ChatGPT, this includes the steps: pre-training, fine-tuning, and reward learning. For this purpose we rely on the pre-trained GPT models, and use our compiled dataset of generated and annotated texts (as explained above) for further reward learning and fine-tuning. 
The reward function used to fine-tune InstructGPT is trained using a dataset of pairs of generated texts with a human-labelled judgement on which text is better, with the objective to maximise the score difference between the `winning' and `losing' texts. The purpose of collecting coarse human labelling is to mitigate any mismatch between the true objective and the preferences of human annotators, thus increasing inter-annotator agreement~\cite{stiennon2022learning}. Nevertheless, relying solely on general annotations, as in the case of InstructGPT, results in a reward function that fails to shed light on the quality of texts across various aspects, making it too broad to apply in narrower tasks and fields. To address this limitation, we can take advantage of the existing high-quality annotations available to us from skilled and experienced professional human annotators. By exploiting their expertise, we can train a more nuanced reward function that offers fine-grained evaluation and provides interpretable scores, aligning more effectively with our specific research goals.
%\textcolor{blue}{
Finally, we can evaluate different methods of content generation on our reading practice platform, Read\&Improve\footnote{\url{https://readandimprove.englishlanguageitutoring.com/}} \cite{randi}. By collecting both implicit user feedback -- which texts they engage with more by spending longer reading them, clicking on definitions, completing the tasks, \emph{etc} -- and explicit user feedback (e.g.\ by asking them to rate texts and express opinions) we can assess which LLM-driven systems are most successful.
%}

\section{Calibrating Assessment Items and Teaching Materials}

In addition to content creation, LLMs can potentially be leveraged for the evaluation and calibration of existing learning content: test items and teaching content.
An example of this is question difficulty estimation (QDE) from text, which has received increasing research interest in recent years \cite{benedetto2023survey,alkhuzaey2021systematic}. QDE from text offers a way to overcome the limitations of traditional approaches such as manual calibration and statistical analysis from pre-testing. These traditional approaches are either subjective or introduce a long delay between item creation and deployment due to the complexities of pre-testing on sizeable and representative populations.

QDE from text is a regression task, where the model is asked to provide, given the text of the question, a numerical estimation of its difficulty on a given scale.
It can be either a supervised or unsupervised task, depending on whether a dataset of already calibrated exam questions is available, and LLMs have been used in both scenarios.
Regarding supervised estimation, LLMs that leverage transfer learning -- specifically, BERT \cite{bert} and DistilBERT \cite{distilbert} -- are the current state of the art \cite{zhou2020multi,benedetto2021application} and have been shown to outperform other approaches using traditional NLP-derived features \cite{benedetto2023quantitative}.
There has been less work on unsupervised estimation but, even in this scenario, LLMs have been shown to be helpful for estimating question difficulty from text \cite{loginova2021towards}.

All the models proposed in previous research require some kind of transfer learning starting from the publicly available pre-trained models, which might be expensive and not feasible in all scenarios.
Bigger LLMs, such as the aforementioned GPT models, could be used for zero-shot or few-shot difficulty estimation from text, which is yet to be explored. 
As an example, ChatGPT can be asked to rank given questions by difficulty, and it can also provide an indication of the specific difficulty level (e.g.\ \textit{Easy}, \textit{Medium}, \textit{Hard}).
Crucially, the difficulty of a pool of questions depends on the specific student population that is assessed with them, and it is difficult to provide the LLM with all the information required to describe the specific pool of learners that will be assessed with the items.
The model seems to be -- at least partially -- capable of distinguishing between different CEFR levels, since the same question can be assigned different levels depending on whether the model is asked to consider learners of level A1 or C1.
However, extensive experiments should be carried out to better evaluate this, as the model sometimes performs counterintuitive estimations: in our preliminary experiments, for instance, ChatGPT sometimes estimated a question to be more difficult for C1-level learners (i.e., ``Proficient'') than A1-level learners (i.e., ``Basic'').

\section{Automated Assessment of Language Learners}

Automated assessment has long been a prominent task in educational applications research: for instance assessing learner English speech \cite{speechrater,knill2018impact,zechner-evanini,craighead-etal-2020-investigating} and writing \cite{Burstein_Chodorow_Leacock_2004,andersen-etal-2013-developing,alikaniotis-etal-2016-automatic,riordan-etal-2017-investigating,andersen-etal-2021}. Here we focus on writing and the task of `automated essay scoring' (AES).
Whereas previous systems have involved feature engineering -- typically centred around informative sequences of characters, words, part-of-speech tags, as well as phrase structures from a parser and automatically detected errors \cite{andersen-etal-2013-developing,yannakoudakis2018developing} -- more recent research systems have involved neural models for assessment \cite{alikaniotis-etal-2016-automatic,riordan-etal-2017-investigating,andersen-etal-2021}. These models tend to be carefully crafted and evaluated, since language assessment can be a task with major consequences for the learner, including education and career prospects. Therefore any involvement of LLMs in assessment systems must be approached cautiously and its impact measured on existing benchmarks. Deployment of LLM-based assessment models should be restricted to human-in-the-loop low-stakes contexts first, including practice applications \cite{yannakoudakis2018developing,randi} or placement tests such as Linguaskill\footnote{\url{https://www.cambridgeenglish.org/exams-and-tests/linguaskill/}}.
%or the Duolingo English Test\footnote{\url{https://englishtest.duolingo.com/applicants}}.

The idea of using ChatGPT for assessing students' answers was put forward by Jeon \& Lee \cite{jeon2023large}.
Further practical steps were taken by Mizumoto \& Eguchi \cite{mizumoto2023exploring} who experimented with GPT-3.5 for AES on 12,000 essays from the ETS Corpus of Non-Native Written English (TOEFL11) \cite{blanchard2013toefl11}, compared the scores to benchmark levels on a 0-9 scale, and concluded that a GPT-only model only achieves weak agreement with the reference scores ($.388$ quadratic weighted kappa).
The authors furthermore compare the GPT scorer with several models involving various combinations of 45 linguistic features -- related to lexical diversity, lexical sophistication, syntactic complexity and dependency, and cohesion -- and observe that although the GPT baseline is outperformed by the linguistic features on their own, the best results are obtained by combining the two approaches ($.605$ QWK).
This is a finding similar to previous research, as the previous state-of-the-art performance was obtained by combining BERT-style neural models and feature-based models \cite{andersen-etal-2021,lagakis2021automated}.

One potential use for LLMs regarding assessment that, to the best of our knowledge, has not been thoroughly explored is for explaining assessment predictions. \emph{Explainable AI} is an emerging research topic, a regulatory prospect \cite{ec}, and a challenge for NLP models dependent on `black box' neural networks. One possibility is to adopt the `chain-of-thought' prompting style \cite{wei2022chain} -- as opposed to zero-shot or few-shot prompting \cite{brown-gpt3} -- to elicit explanations about assessment decisions from LLMs. Exactly how to engineer a series of prompts for the LLM in the chain-of-thought style is a matter for investigation, but for instance they could be similar to the following (albeit longer to elicit better explanations):

\begin{verbatim}
    A class of students has been given the essay prompt: <essay_prompt>.
    This student essay -- <example_text>
    -- was given a score of <example_score>.
    Explanation: the use of language and grammatical resource are advanced
    but there is a spelling error in the first sentence and a grammatical 
    error in the final sentence.
    This student essay -- <target_text>
    -- has been given a score of <predicted_score>.
    Please give an explanation why the text was given this score.
\end{verbatim}

\noindent
The aim of such an approach would be to obtain explanations specific to the essay, pinpointing relevant sections from the text if possible, and grounded in marking criteria for the learner's target level so that the explanation is relevant and useful.
In common with other tasks described in this paper, further research with LLMs and proper evaluation on existing benchmarks and by human experts is needed before we can definitively conclude that this is a research avenue worth exploring.

Finally we note that there are concerns around LLMs being used in fraudulent ways by learners, but that plagiarism concerns are long-standing in computer-based exam settings. If proctoring software is set up to prevent text import from elsewhere (e.g.\ disabling copy-and-paste keyboard shortcuts) or to detect bursty text insertion through keystroke logging, then this is one defence against exam malpractice from LLM text generation or any other online source. In this way, LLMs are an extension of a threat we are already familiar with. Furthermore, automatic detection of LLM-generated text is the subject of the AuTexTification (Automated Text Identification), part of IberLEF 2023 (the 5th Workshop on Iberian Languages Evaluation Forum) co-located with the SEPLN Conference this year\footnote{\url{https://sites.google.com/view/autextification}}. It appears that the level of performance from submitted systems has been high, outdoing a logistic regression baseline in many cases, with system descriptions to be presented in September. It may be that such systems can be employed as an additional line of defence against exam malpractice involving LLM-generated text.

\section{Providing Feedback to Language Learners}

As a precursor to providing automatic lexico-syntactic feedback to language learners, one requirement is to first carry out the NLP task of \emph{grammatical error detection} (GED) or \emph{grammatical error correction} (GEC). These tasks have a rich pedigree involving various benchmark corpora \cite{yannakoudakis-etal-2011-new,boyd2018using,naplava2019grammatical,syvokon-etal-2023-ua}, research papers \cite{dahlmeier2012beam,bell-etal-2019-context,caines-etal-2020-grammatical,yuan-etal-2021-multi,qorib-etal-2022-frustratingly} and shared tasks \cite{ng-etal-2013-conll,ng-etal-2014-conll,bryant-etal-2019-bea} -- all of which enable us to establish that the current state-of-the-art approach to GED and GEC tends to involve supervised fine-tuning of neural network language models using carefully annotated training data \cite{bryant-et-al-2023}.
The recent emergence of LLMs, however, offers the prospect of developing GED or GEC models which are largely unsupervised other than some examples for few-shot learning\footnote{Note that LLM training is often described as `self-supervised' due to the human-authored training data, but for the purpose of GED/GEC, we say `unsupervised' because in this context no task-specific training data is required.}.
%While language models have for a long time been put to use for the NLP task of grammatical error correction (GEC) \cite{dahlmeier2012beam}, the recent emergence of LLMs present a renewed opportunity to explore unsupervised techniques to correct grammatical errors and provide informative feedback for students and teachers in the language learning domain.

The challenge ahead, in common with the application of LLMs to other tasks described in this paper, is to properly benchmark LLM-based models for GED and GEC on existing corpora, so that their performance can be compared to previous models.
Some preliminary work has been done towards this aim, as described in a recent survey of GEC \cite{bryant-et-al-2023}.
For instance, Wu \emph{et al.} \cite{wu2023chatgpt} and Coyne \& Sakaguchi \cite{coyne2023analysis} present preliminary results applying LLMs to GEC. The former compares ChatGPT to Grammarly and GECToR \cite{omelianchuk2020gector}, a previous state-of-the-art GEC system, and the latter compares GPT-3.5\footnote{Specifically, they use text-davinci-003.} to two other GEC systems \cite{yasunaga2021lm, liu2021neural}. Both approaches find the GPT* models perform worse than existing systems when measured using automatic evaluation techniques on existing benchmark corpora (namely CoNLL-2014 \cite{ng-etal-2014-conll}, JFLEG \cite{jfleg}, BEA-2019 \cite{bryant-etal-2019-bea}). The authors ascribe this to the model's tendency to \textit{over-correct} learner text; by inserting additional text or re-structuring phrases, the corrected text moves further from the original text and is penalised by the automatic scorers. However, both works carry out human evaluation to rate the output from each system and find a preference for the GPT* output because the corrected sentences tend to be more fluent. At the same time, they found instances of \emph{under}-correction in the human-generated reference sentence: in other words GPT* models catching and correcting errors which were not corrected by the expert annotators.

While both approaches are preliminary and human evaluation tentative -- based on only small samples of 100 sentences at a time from each test set -- overly fluent corrections present a challenge for automatic evaluation methods as they are much more open-ended than minimal edits targeting grammatical errors rather than stylistic choices. Furthermore, while fluent corrections may at times be preferred by human evaluators, they may not aid language learners if they drift too far from the original text. Existing annotation guides for error correction state that edits should be as minimal as possible so that the learner can be helped to express what they are trying to say, rather than told how to express it differently: that is, how to amend an error rather than avoid it \cite{nicholls-2003}. The issue is not a new one \cite{sakaguchi-etal-2016-reassessing} but remains a matter for further investigation under the new conditions presented by more capable LLMs.

Another potential use of LLMs in this area is providing automatically-generated feedback comments to learners to explain linguistic concepts, grammatical points or semantic nuance. Indeed there was a recent shared task on \emph{feedback comment generation} \cite{nagata-etal-2021-shared} where, when presented with an erroneous sentence such as, ``He agrees the opinion'', the task was to produce a comment such as: \texttt{The verb agree is an intransitive verb and cannot take direct objects: add the appropriate preposition}\footnote{\url{https://fcg.sharedtask.org/}}. Participants in the shared task were able to outperform the baseline system (`an encoder-decoder with a copy mechanism based on a pointer generator network'\footnote{\url{https://github.com/k-hanawa/fcg_genchal2022_baseline}}) through careful feature extraction from parsers and GEC models, combined with prominent LLMs at the time such as T5 or GPT-Neo \cite{gpt-neo} (e.g.\ Babakov \emph{et al.} \cite{babakov} achieved second place in the shared task; developers of the first-placed entry have not published a system description to the best of our knowledge). It remains to be seen whether current LLMs can be tuned towards even better performance on this task: it may be that the pre-processing of texts to obtain additional linguistic information and the incorporation of pre-defined templates will continue to be vital for accurate and sensible feedback comment generation, even with ever-larger LLMs involved \cite{coyne-2023-template}.
%\textcolor{blue}{
These are methods we can trial through A/B testing of different feedback models on our essay-writing practice platform, Write\&Improve\footnote{\url{https://writeandimprove.com/}} \cite{andersen-etal-2013-developing,yannakoudakis2018developing}.
%}

Other applications of LLMs for language learning feedback include chatbot interaction to explain linguistic concepts -- akin to the `Explain My Answer' feature in Duolingo Max, but also going beyond this with dialogue which is adaptive to the learner level \cite{tyen-etal-2022-towards} -- word suggestion, paraphrasing and translation to aid learners with essay writing, and document-level feedback on, for instance, inter-sentence coherence markers, co-reference and anaphoric reference, maintaining tense and aspect consistently, argumentation structure, task completion and more. Key desiderata are that the feedback should be accurate, based on evidence, personalised, inoffensive and preferably linked to teaching materials so that the learner may continue to benefit from EdTech applications for language.

\section{Risks \& Ethical Considerations}

We advocate for a cautious approach to the incorporation of LLMs in EdTech for language learning, in which the training process, performance and limitations, and pathway to delivery are well documented and the risks of misapplication of such technology are understood. There are general concerns about AI for NLP and education which are recorded in the literature and continue to be relevant, perhaps more so, as LLMs come to the fore.
Firstly there is a bias towards English, and specific genres of English, due to a combination of commercial pressures, training data availability, and data sourcing from the World Wide Web: even though several models have been trained in multilingual ways, the general trend with LLMs has exacerbated this pre-existing bias \cite{gururangan-etal-2022-whose,sogaard-2022-ban,ramesh-etal-2023-fairness}.
As LLMs grow, so does their climate impact: an issue which interacts with societal and infrastructure complexities but which we should nevertheless bear in mind and attempt to mitigate \cite{strubell-etal-2019-energy,dodge2022}.
In addition, LLMs are known to exhibit certain biases \cite{barocas-et-al} -- both \emph{representational} (language use around demographic groups) \cite{blodgett-etal-2020-language,stochastic-parrots} and \emph{allocational} (how a system distributes resources or opportunities) \cite{suresh-guttag,blodgett-etal-2020-language} -- which need to be debiased or otherwise controlled \cite{kaneko-bollegala-2021-debiasing,lalor-etal-2022-benchmarking}.

Suresh \& Guttag identified various sources of harm in the `machine learning life cycle' \cite{suresh-guttag}: historical bias, representation bias, measurement bias, learning bias, aggregation bias, evaluation bias, deployment bias. They note that effects cascade downstream and cycle around ML systems. They provide some mitigation strategies and reference previous work in this area \cite{friedler-et-al,finocchiaro-et-al}. Kasneci \emph{et al.} \cite{kasneci2023chatgpt} also point to copyright issues with output from LLMs which are largely unresolved, as well as concerns about pedagogical and learning effects: namely that both teachers and students, ``may rely too heavily on the model'', and that it may be difficult to, ``distinguish model-generated from student-generated answers'' \cite{dehouche2021plagiarism,cotton-et-al,kirchenbauer2023watermark}. In addition they raise data privacy and security issues which require firmer regulation and auditing of EdTech firms, the problem of false information issued by LLMs, and of designing appropriate application interfaces which are both engaging and beneficial to end-users. 
It is worth noting that NLP researchers have made some attempts at using LLMs to assess the trustworthiness of generated texts, which could go some way towards mitigating the false information problem \cite{manakul2023selfcheckgpt,lee2020language,peng2023check}.

Regarding AIED and language learning, LLMs present specific risks relating to generated outputs which may be inaccurate, confusing, offensive, and so on -- risks which are present in human teachers too, but made no less harmful as a result. For this reason the most successful systems may be human-machine hybrids, with humans in-the-loop or similar, where LLMs are viewed as assistive technology for human experts rather than replacements for them -- performing the more mundane and mechanical tasks while experts provide the inputs characteristic of human interaction \cite{santoro-monin}. Another way that humans can monitor LLM outputs is through evaluation, and feedback mechanisms for systems in production, so that problematic outputs may be flagged.

We can also look at standards for `responsible AI' published by technology firms and research institutes \cite{prabhakaran2022human,thakkar-et-al-2022,khan-hanna-2023}\footnote{\url{https://huggingface.co/blog/ethical-charter-multimodal}}. For example, Duolingo \cite{burstein-et-al-2023} sets out its approach to responsible AI under Validity \& Reliability, Fairness, Privacy \& Security, Accountability \& Transparency -- all of which have been touched on in this paper. Regarding the last attribute in particular -- Transparency -- it is apparent from recent media stories that more can be done in this area in terms of educating the general public about how LLMs are trained, how trustworthy they may or may not be, and how best to interact with them. This is a general problem but one which nonetheless presents a challenge for EdTech applications.

\section{Conclusion}

In this paper, we have explored the opportunities for language-learning EdTech offered by `generative AI' through LLMs. 
We conclude that preliminary indications are promising, but that the best systems may still require human intervention and/or the inclusion of well-established linguistic features.
It may well be that LLMs can enhance language-learning EdTech, if we can establish the following through further empirical work:

\begin{enumerate}
\itemsep0em
\item that models enhanced by LLMs perform better than existing models on established benchmarks, or on alternative evaluation metrics which need to be defined in order to properly probe LLM capabilities for language teaching and assessment \cite{srivastava2022imitation} -- moreover that performance is \emph{sufficiently} better to justify the additional costs in computing and environmental terms;
\item that LLM-enhanced technology is of benefit to language learners, whether that is measured through engagement, enjoyment, learning outcomes or some combination of the three;
\item that LLM-enhanced technology does not disadvantage relevant groups (learners, teachers, writers and editors of materials, examiners) whether through bias, misinformation, or adversely affecting student progress
%diminished employment prospects
-- instead, the technology should be assistive to all groups in some regard.
\end{enumerate}

\noindent
Finally, we note that LLMs should not be over-hyped as an AI revolution, but rather as an evolutionary step in neural network models -- the inevitable result of the inexorable growth in network size since the Transformer was first applied to language tasks in 2017 \cite{vaswani}.
LLMs represent a milestone on an evolutionary path which has been unfolding for many years and thus is well documented in open access publications and open source code repositories. If we maintain this tradition -- by close inspection of proprietary models, or opting to use models trained in open ways -- it will be of benefit both to future researchers and scientific development, but also users of AI applications who require some transparency regarding the technology.
Harmful bias and other risks remain an ongoing challenge for developers of AI systems, and LLMs deployed in language learning EdTech may only exacerbate these. Therefore, proper mitigations should be put in place to address the issues which have been identified in this paper and elsewhere.

Nevertheless, LLMs present a great opportunity to continue improving EdTech for language learning, including novel ways to generate content, provide feedback, and deal with other linguistic features which hitherto have not been commonly attempted: for instance, chatting in open-ended ways at the level of the learner \cite{tyen-etal-2022-towards}, providing document-level assessment and feedback \cite{wambsganss-etal-2022-alen}, handling code-switching or `plurilingual' learning \cite{nguyen2022building}.


\begin{acknowledgments}

This work was supported by Cambridge University Press \& Assessment.
We thank Dr Nick Saville and Professor Michael McCarthy for their support.
We are grateful to the anonymous reviewers for their helpful comments.

\end{acknowledgments}

%%
%% Define the bibliography file to be used
\bibliography{bib}

\end{document}